\documentclass{article}

\usepackage[preprint]{corl_2024} 

\usepackage{times}

\usepackage{natbib}
\usepackage{multicol}
\usepackage[normalem]{ulem}
\usepackage{color}
\usepackage{xcolor}

\newenvironment{flushitemize}{%
\begin{list}{$\bullet$}
   {\setlength{\leftmargin}{15pt}}%
    \setlength{\labelwidth}{20pt}
    \setlength{\itemindent}{0pt}
    \setlength{\labelsep}{0.5em}
 \setlength{\itemsep}{1pt}
 \setlength{\parskip}{0pt}
 \setlength{\parsep}{0pt}}
 {\end{list}}
 
\usepackage{outlines}
\usepackage{caption}
\usepackage{graphicx}
\usepackage{subcaption}
\usepackage{amsmath,amssymb}
\usepackage{wrapfig}

\usepackage{amsthm}
\newtheorem{example}{Example}

\title{Robot Learning as an Empirical Science: \\ Best Practices for Policy Evaluation }

\author{
  Hadas Kress-Gazit$^{1,2}$, Kunimatsu Hashimoto$^2$, Naveen Kuppuswamy$^2$,  Paarth Shah$^2$,\\ \textbf{Phoebe Horgan$^2$, Gordon Richardson$^2$,
  Siyuan Feng$^2$, Benjamin Burchfiel$^2$} \\
  $^1$Cornell University and $^2$Toyota Research Institute\\
  Corresponding author: \texttt{hadaskg@cornell.edu} 
}

\begin{document}
\maketitle


\begin{abstract}
The robot learning community has made great strides in recent years, proposing new architectures and showcasing impressive new capabilities; however, the dominant metric used in the literature, especially for physical experiments, is ``success rate'', i.e. the percentage of runs that were successful. Furthermore, it is common for papers to report this number with little to no information regarding the number of runs, the initial conditions, and the success criteria, little to no narrative description of the behaviors and failures observed, and little to no statistical analysis of the findings. In this paper we argue that to move the field forward, researchers should provide a nuanced evaluation of their methods, especially when evaluating and comparing learned policies on physical robots. To do so, we propose best practices for future evaluations: explicitly reporting the experimental conditions, evaluating several metrics designed to complement success rate, conducting statistical analysis, and adding a qualitative description of failures modes. We illustrate these through an evaluation on physical robots of several learned policies for manipulation tasks.
\end{abstract}


\section{Introduction}
%

Recent years have seen significant advancements in machine learning, with successful deployments of machine learning models in the wild now becoming commonplace \cite{SAM, stablediffusion, gpt4}. Robotics has also undergone significant changes with increasing adoption of data-driven machine-learning methods. These range from reinforcement learning (RL)~\cite{RLsurvey}, to deep RL~\cite{DeepRLsurvey} to the recent emergence of foundation models - general-purpose perception-action models \cite{octo_2023, rt12022arxiv, brohan2023rt2} trained on large and diverse datasets and capable of performing in myriad in-the-wild domains. 

As the field progresses, 
complex behaviors have been shown both in simulation and with physical hardware. However, while papers typically describe their architectural designs and training paradigms in depth, their evaluation criteria and process are often sparse in details that would help move the field forward. Specifically, 
the evaluation often solely focuses on ``success rate'' --the percentage of autonomous runs that were successful-- with little description of the experimental conditions, number of evaluations, success criteria, performance, failure modes, and typically without any statistical analysis. This lack of detail and nuance makes it difficult to assess the true state of the field and impacts two communities of researchers; those who develop learning algorithms, and those who wish to use them. The former because it is not clear what the exact state of the art is and what the fruitful research directions are. The latter, who may wish to use policies as a black-box in some larger system, because they do not have a clear understanding of possible failure modes and under what conditions the algorithms were evaluated. 

\textbf{Contribution}: We propose best practices for policy evaluation to improve the science of robot learning. These include suggestions for the experimental setup, different metrics, and the analysis of the results. While we focus on evaluation with physical robots, these best practices extend well to simulation. We illustrate our points through several examples of manipulation tasks performed by physical robots.    

\begin{figure*}
     \centering
     \begin{subfigure}[b]{0.3\columnwidth}
         \centering        \includegraphics[width=\columnwidth]{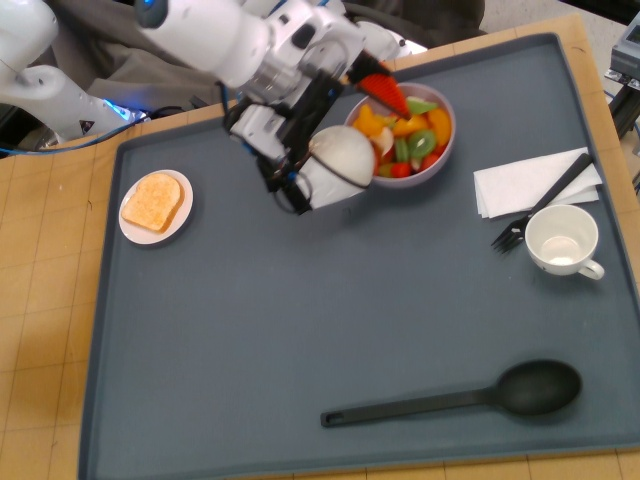}
         \caption{\textit{Bowl}}
         \label{fig:BowlTask}
     \end{subfigure}
     \begin{subfigure}[b]{0.3\columnwidth}
         \centering         \includegraphics[width=\columnwidth]{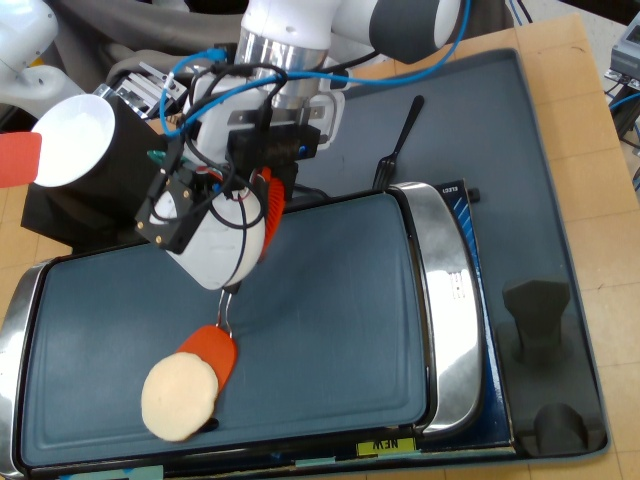}
         \caption{\textit{Pancake}}
         \label{fig:PancakeTask}
     \end{subfigure}
     \begin{subfigure}[b]{0.35\columnwidth}
         \centering         \includegraphics[width=\columnwidth]{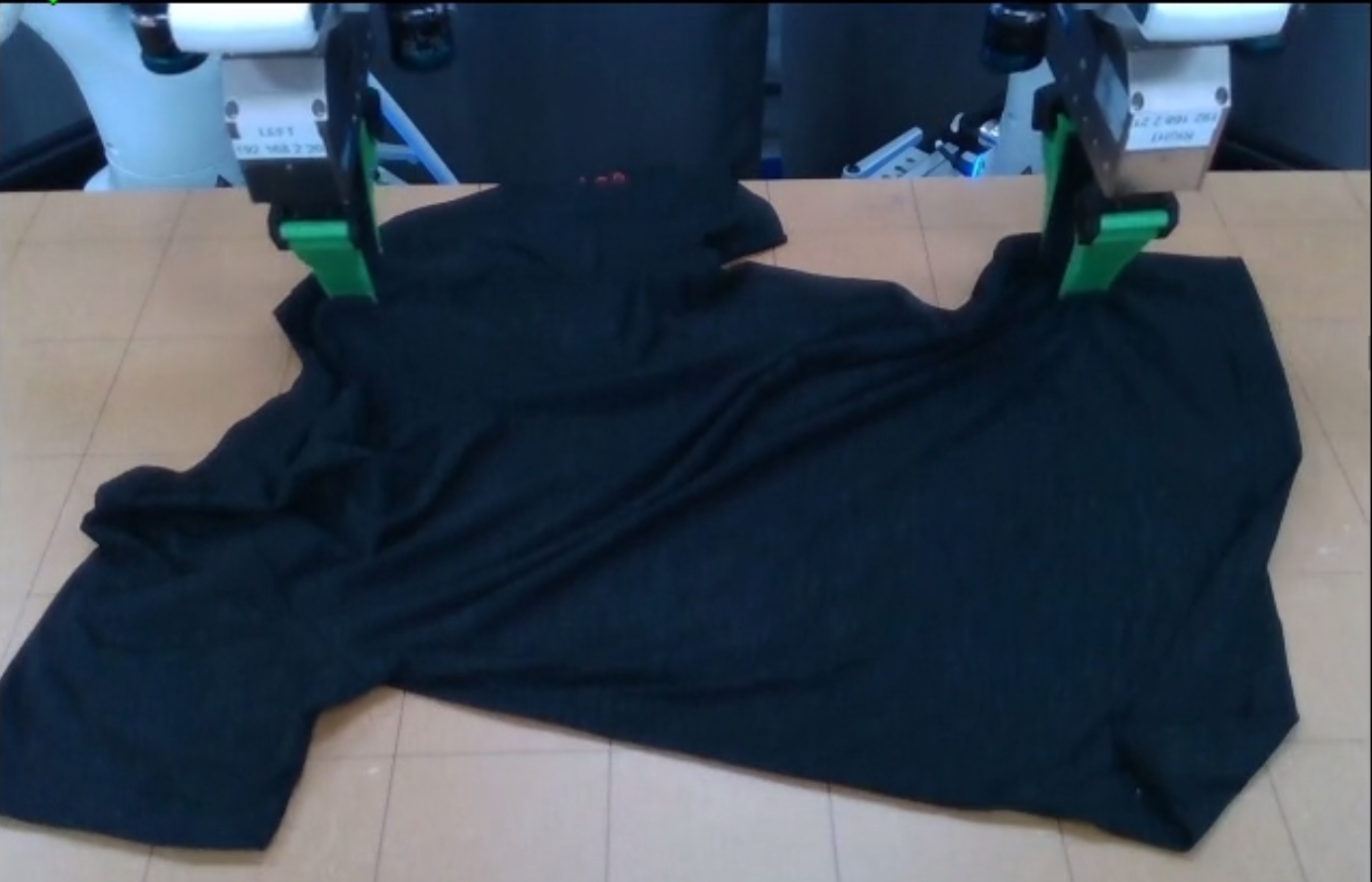}
         \caption{\textit{Shirt}}
         \label{fig:ShirtTask}
     \end{subfigure}
        \caption{Tasks and robots }
        \vspace{-1.2em}
        \label{fig:tasks}
\end{figure*}

\textbf{Data used in this paper:} We illustrate our proposed best practices using data collected from physical robot evaluation runs. For each skill, we train several behavior cloning (BC) policies; all policies use the same set of human teleoperated demonstrations, but differ in 
 architecture, observation space, and hyper parameters. While this paper considers a set of learned robot policies as case studies, we treat their implementation details as black-boxes.
\begin{flushitemize}
    \item \textbf{Push Bowl} (\textit{Bowl}, Fig.~\ref{fig:BowlTask}, 6 policies): Starting with 
    the end effector is in the air, a Franka Emika Panda arm pushes a bowl filled with (fake) fruit to a specific area on the table. 
    \item \textbf{Flip and Serve Pancake} (\textit{Pancake}, Fig.~\ref{fig:PancakeTask}, 3 policies): Two Franka Emika Panda arms perform the task together; first each arm grasps a spatula, then the right arm flips the pancake, then the  left arm lifts the pancake and places it on a plate. 
    \item \textbf{Fold Shirt} (\textit{Shirt}, Fig.~\ref{fig:ShirtTask}, 2 policies): Two Franka Emika Panda arms perform the task together; they grasp a T-shirt, folds it three times, and center it on the table. 
\end{flushitemize}
\subsection{Related Work}
\noindent\textbf{Experiment design and analysis in robotics and machine learning:} As disciplines mature, they begin forming best practices for evaluation and reporting, to ensure further progress. Such guidelines help communicate what information is useful to the community and set expectations, which is especially useful for people entering the field. Examples for best practices in fields that are close to robot learning include a primer on conducting experiments in human-robot interaction~\cite{Hoffman2020}, and statistical analysis for evaluation of deep reinforcement learning~\cite{agarwal2021deep,Henderson_Islam_Bachman_Pineau_Precup_Meger_2018}.

\noindent\textbf{Metrics in robotics and related fields:}
The most common metric used in the robot learning literature 
is ``success rate'' (e.g.~\cite{brohan2023rt2, bharadhwaj2023roboagent,open_x_embodiment_rt_x_2023}). This is true both for simulation and physical experiments; however due to the fact that in simulation we have the full state of the world, there are more explicitly defined success metrics for simulated tasks. One common metric is distance from goal, especially in pick and place type tasks~\cite{yu2021metaworld,gupta2019relayFrankaKitchen,Rajeswaran-RSS-18}.
For physical experiments, especially work that evaluates different tasks, some works do not state explicitly what the criteria is, and others rely on human judgement  (e.g.~\cite{ma2023vip,saycan2022arxiv}). Recent work has proposed a metric for behavior entropy, capturing a model's ability to create diverse behaviors~\cite{jia2024diverse}. For more specific tasks, researchers have created quantitative metrics such as weight of material transferred during a scooping and a pouring task~\cite{zhou2023TOTO}, time to fall, angle of rotation, torque applied for an in-hand manipulation task~\cite{qi2022hand}, and end state belonging to a predetermined set of states, evaluated manually~\cite{bousmalis2023robocat}.
Several works also capture sub-goal achievement, as we discuss in this paper. Some of these are in simulation where it is possible to automate sub-goal evaluation, especially if it is related to location or contact in space, and some manually evaluated for physical experiments (e.g.~\cite{Heo-RSS-23FurnitureBench,ajay2023compositional}). 
More broadly, metrics have been proposed in adjacent engineering fields including: human-robot interaction studies~\cite{HRImetrics}, natural language processing~\cite{nlpMetrics}, and computer vision~\cite{russakovsky2015imagenet,salimans2016improved,borgefors1984distance, rubner2000earth}.

\section{Experiment Setup}
Ultimately, the purpose of an experimental evaluation in the field of robot learning is to gain knowledge and insight about design choices of the learned policy (training, architecture, learning objective, hyper parameters, etc.). To gain the insight, policies would ideally be compared under identical conditions; however, as opposed to simulation, with a physical system it is impossible to replicate the exact same experimental conditions between evaluation runs. Furthermore, we know that the performance of learned policies is highly dependent on the experimental conditions. In this section, we discuss experiment best practices to mitigate this challenge and maximize the utility of valuable and expensive physical experiments.  When employed, they create a (closer to) level playing field for the learned policies being compared, reduce confounding variables that decrease the signal-to-noise ratio of experiments, and allow for increased confidence in experimentally-drawn conclusions. 

\subsection{Success criteria}
First and foremost, there must be a clear, detailed, and unambiguous definition of success. While this sounds obvious, many papers do not provide a definition of success. When success criteria are not explicit, evaluation may become biased; as an anecdote, we note a time when three of the authors watched a robot pour ice from a cup into a sink. The robot did pour the ice into the sink but then hesitated when lowering the cup, moving it for a few seconds above the table, ultimately placing it on the table and tipping it over. Two of the authors viewed the episode as a failure, while the third thought it was a success. This ambiguity allows inadvertent evaluation bias to distort the results. 


\subsection{Initial conditions}

It is well known among robot learning practitioners that today's learning-based robots are highly sensitive to their deployment environment. For non-interactive tasks, the most critical aspect of this are the initial conditions from which rollouts are performed - these include object types and locations in the environment, lighting conditions and camera locations, to name a few.

It is easy and commonplace to inadvertently change initial distributions during evaluation (e.g. object placements, environmental shifts such as sunlight, background changes) in ways that could significantly affect policy performance but are not immediately obvious to a person. Unfortunately, it is not common for empirical work to control for these details or explicitly describe the initial conditions used across evaluation runs. 
\begin{figure*}
     \centering
     \begin{subfigure}[b]{0.45\columnwidth}
         \centering        \includegraphics[width=\columnwidth]{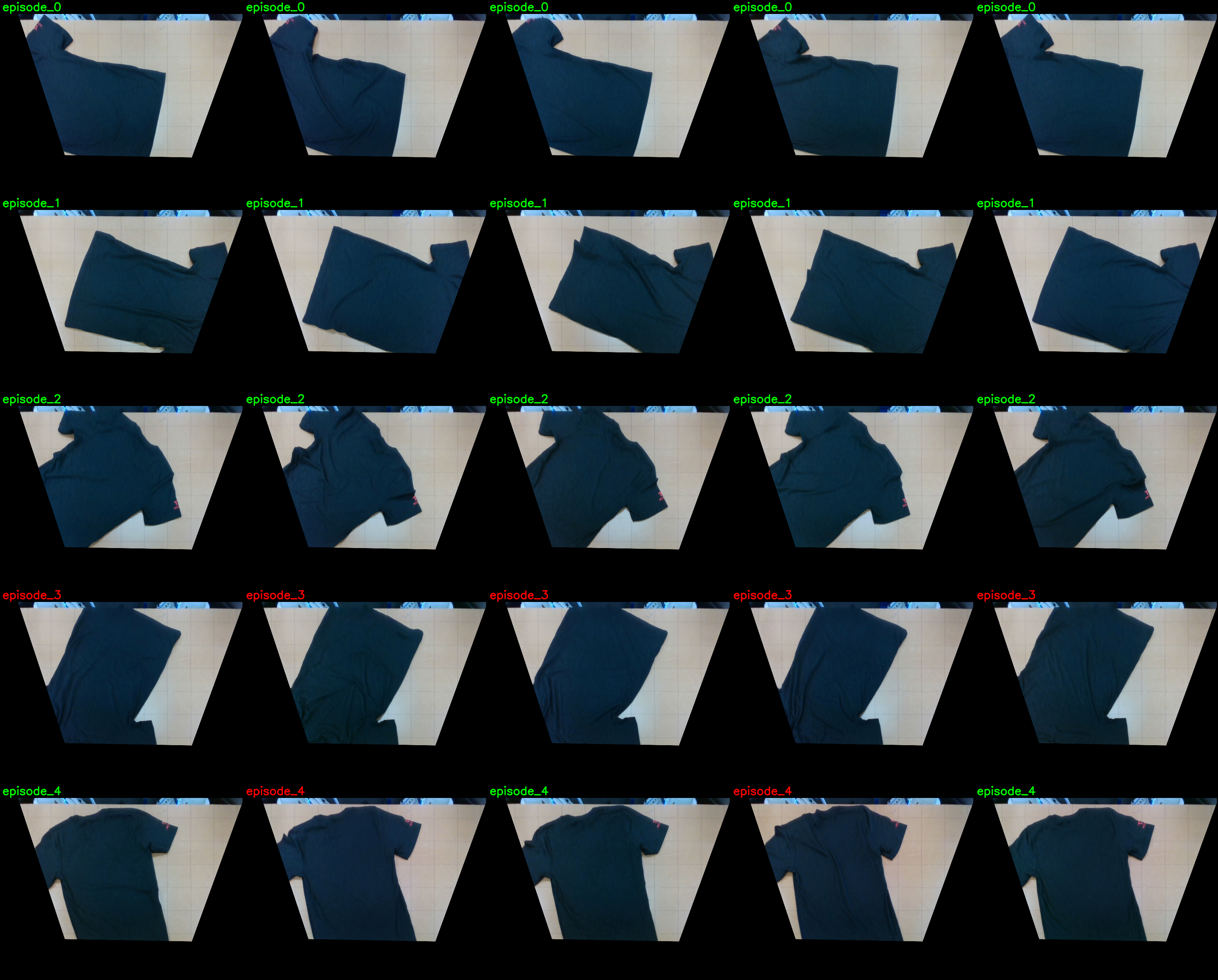}
         \caption{\textit{Policy A}}
         \label{fig:STL}
     \end{subfigure}
     \begin{subfigure}[b]{0.45\columnwidth}
         \centering         \includegraphics[width=\columnwidth]{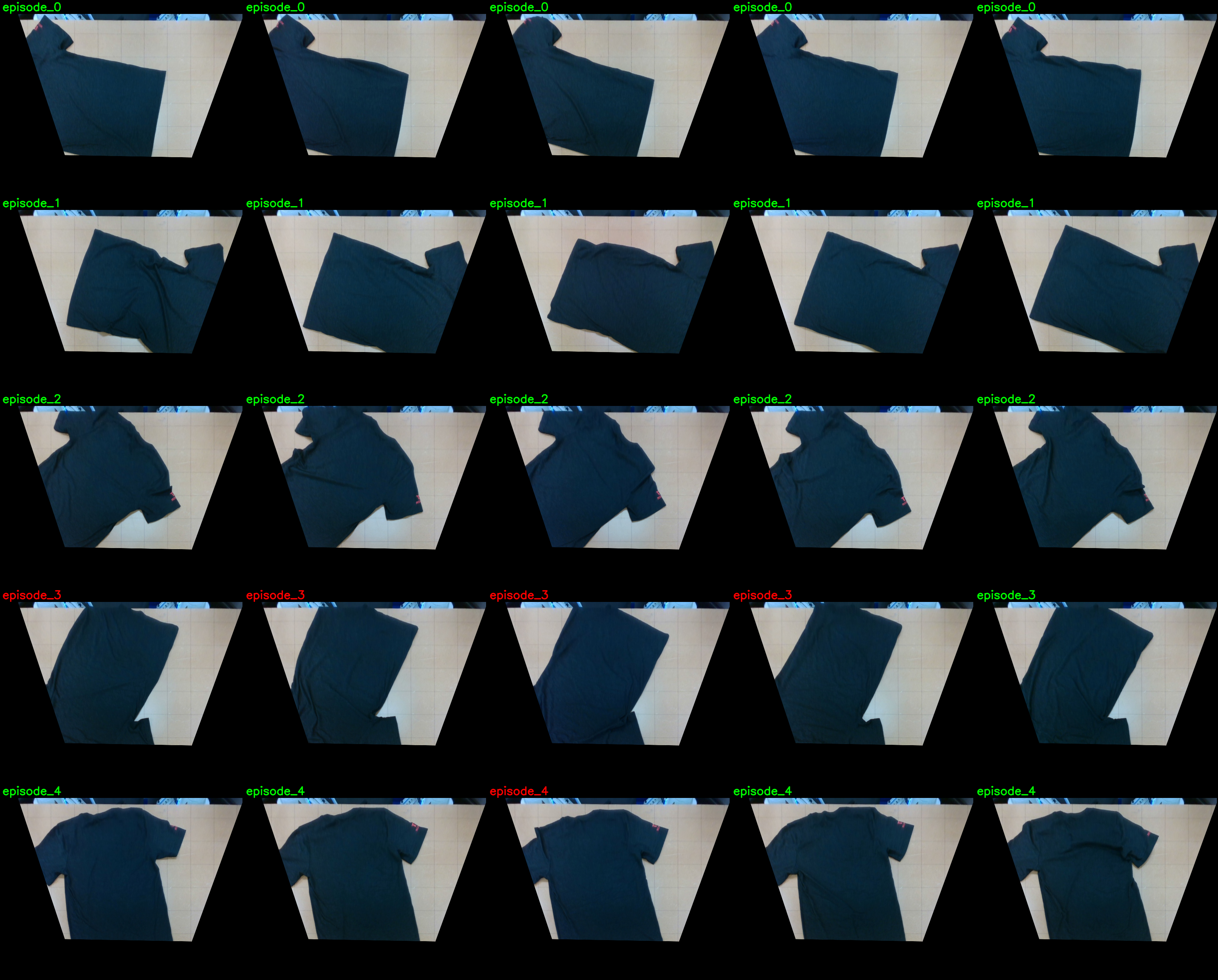}
         \caption{\textit{Policy B}}
         \label{fig:shirtIC}
     \end{subfigure}
        \caption{Initial conditions (ICs) for Example~\ref{exm:init}; Policy A on the left and Policy B on the right. Each row represents 5 evaluations with the same IC; the ICs are numbered 0-4 from the top row. The color of the text above each photo indicates the success; green is a successful rollout, red is a failure. We can see that the IC are visually consistent. Furthermore, for ICs 0,1,2 both policies always succeed, while for IC 3 they mostly fail.}
        \label{fig:shirtIC}
        \vspace{-0.5cm}
\end{figure*}

\begin{example}[Effect of initial condition]
\label{exm:init}
Figure~\ref{fig:shirtIC} shows the set of initial conditions (ICs) we used to evaluate policies A and B for \emph{shirt}. While the overall success rate of policy A and B are similar (72\% vs 80\%), we can see that the success is highly dependent on the IC; if we were not careful in controlling for IC, we might arrive at a different conclusion. For example, if we used different ICs for each policy, and by chance use mainly ICs 1,2 for one policy and ICs 3,4 for the other, we may conclude that one policy is significantly more performant than the other. 
\end{example}
\begin{example}[Detecting distribution shift by matching initial conditions]
\label{exm:distDrift}
Going back to Example~\ref{exm:init}, we performed an additional evaluation of the policies, using the same initial conditions, after a lab reorganization. We ran 8 evaluation runs of policy A, 2 each on IC 0,1,3,4 and got 0/8 (0\%) successes. This is in contrast to the previous performance of A on the same initial conditions which was 13/20 (65\%).  By controlling for the initial conditions, we can detect a suspected distribution shift resulting in degradation of the policies. 
\end{example}
\subsection{Experimental Process Best Practices}
In many robot learning papers that describe new algorithmic approaches, such as new architectures, researchers perform ablation studies and the analysis of the new approach is in comparison to the ablations and other baseline approaches. For this comparison to be meaningful, and more importantly, to reduce as much as possible unintended bias in the evaluation, we recommend the following: 
\begin{flushitemize}
    \item Defining detailed success criteria ahead of time for overall success and semantic metrics, as described in Section~\ref{sec:semantic} . For example, “The robot picked up the cup with the ice, poured all the ice cubes into the sink, and placed the cup back on the table without tipping it over” is better than “The robot poured the ice into the sink.” We advocate for the behavior designer to be the person writing the criteria, but not the person evaluating - that way it is easier to detect ambiguous or incomplete descriptions.   
    \item Reducing, as much as possible, the unintended variability in environmental conditions between different policy rollouts, especially in the initial conditions. This can be done by matching initial conditions using image overlays or markings in the scene, and ensuring policies are evaluated in the same session so lighting and other environmental conditions are more likely to be the same. 
    \item A/B testing, i.e., interleaving policy rollouts when comparing different policies in a way that is blind to the evaluator. This means evaluating all of the policies within one session, as opposed to different policies in different sessions; this will mitigate unintended bias from the evaluator since they will not know which policy is running. See~\cite{ABtesting} for a compelling argument. 
    \item Ensuring consistency across evaluators and separating the role of demonstrator and evaluator. People who work closely with the robot and gather demonstrations have a better understanding of how to set up the robot and environment for maximum success; separating the roles and ensuring the same person (or group of people) performs all the evaluations will create a more consistent  assessment of the policies. 
\end{flushitemize}

\section{Evaluation metrics}


In this section we describe several  metrics that can provide different nuanced information regarding the behavior of a robot. We divide the discussion into two overall types of metrics: \textit{semantic information}, and \textit{performance metrics}; the former is a binary type of evaluation with Yes/No as the result, capturing ``correctness" of the behavior. The latter provides a set of continuous numbers that can be thought of as a dense reward in a reinforcement learning setting, capturing the ``quality" of the behavior. For both types of metrics, some are task-agnostic, for example trajectory smoothness, while other are task-specific; we argue that both are important for analyzing the behavior of learned models. 

\subsection{Semantic information}
\label{sec:semantic}
Capturing a notion of ``success" or ``failure", these metrics are designed as Yes/No questions;  they include success rate (percentage of ``Yes"), subgoal completion, and failures modes. While these metrics are task specific (even for success rate, one needs to define what success in the task means), we distinguish between metrics that are descriptive (Section~\ref{sec:rubric}) and metrics that are computable, for example those based on logic (Section~\ref{sec:logic}).  
\subsubsection{Rubrics}
\label{sec:rubric}
To measure semantic task progress, which provides a more fine-grained signal for evaluating a policy and allows the evaluator to gather quantitative information regarding failures, we recommend explicitly defining a rubric for each task. Some of the rubric items can be task agnostic, for example ``did the robot exhibit unexpected collisions", but the majority will be task specific. 

This rubric should be filled out by the evaluator during the evaluation. While this adds an extra burden on the evaluator, in practice, when running evaluations on physical robots the evaluator is there making sure the experiment is progressing as intended; we argue that having them also fill out the rubric is worth the time as it can provide useful information.

\begin{example}[Pancake partial success]
\label{exm:pancake}
 Table~\ref{tab:pancake} displays our rubric that the evaluator filled out when rolling out three different policies. Just looking at the overall task success rate makes policy C seems worthless; however, we can see that all policies were able to grasp the spatulas and flip the pancake, but policies B and C struggled with picking the pancake up (last row of Table~\ref{tab:pancake}) and placing it on the plate (overall task success). 
\end{example}

Recently there has been work on automating the detection of such semantic information, and especially detection of failures, through the use of learned models~\cite{inceoglu2023multimodal} or visual-language models~\cite{guan2024task}. While this is a promising direction, as those papers mention, this is not yet a reliable way to automate the assessment of semantic information for policy rollouts on physical systems.

\begin{table}
    \centering
    \begin{tabular}{ |p{4.5cm}||p{2.2cm}|p{2.2cm}|p{2.2cm}|  }
    \hline
    \textbf{Sub-goal} & Policy A (Y/N)& Policy B (Y/N)& Policy C (Y/N)\\
    \hline
         Overall success? & 15 / 3 & 11 / 6 & 4 / \textbf{19} \\
          \hline
         Robot collided with anything?  & 0  / 18 & 2 / 15 & 0 / \textbf{23} \\
          \hline
         Right arm picked up spatula? & 18 / 0 & 17 / 0 & \textbf{23} / 0 \\
          \hline
         Left arm picked up spatula? & 18 / 0 & 17 / 0 & \textbf{23} / 0 \\
          \hline
         Robot flipped pancake? & 18 / 0 & 16  / 1& \textbf{23} / 0 \\
          \hline
         Robot picked up pancake? & 15 / 3 & 12 /5  & 5  / \textbf{18}\\
    \hline
    \end{tabular}
    \caption{Partial rubric for \textit{Pancake} evaluation of three policies. Of note is Policy C; while the overall success rate is low (17\%), it was able to complete part of the task (picking up the spatulas and flipping the pancake) 100\% of the time (as denoted in bold). Looking only at overall success would deem Policy C to be an unsuccessful policy; however, that is not the full picture, as it is able to accomplish more than half the task consistantly.}
    \label{tab:pancake}
    \vspace{-0.5cm}
\end{table}
\subsubsection{Signal Temporal Logic}
\label{sec:logic}
In contrast to rubrics which require a person to evaluate the results, researchers can create functions that automatically produce granular evaluation data from rollout information. 
This can take the form of predicate functions which are functions from the state of the system to $\mathcal{B}=\{True,False\}$. 

Temporal logics~\cite{EMERSON1990} provide syntax and semantics for formulas that are richer than what can be captured by a predicate; for several temporal logics there exists automated evaluation procedures to assess the truth value of formulas. We propose the use of Signal Temporal Logic (STL) and both its Boolean (``Yes/No") and quantitative semantics, also known as the \textit{robustness metric}~\cite{maler2004monitoring,donze2010robust}(Section~\ref{sec:STLRobust}) as a way to create rich, computable metrics. We provide the syntax and semantics of STL in the appendix.

STL formulas can capture properties as simple as ``maintain a distance from an obstacle", and as complex as `` the right arm should be on the right side of the table unless the left arm dropped the spatula, in which case the right arm should pick up the spatula within 10 seconds of it being dropped" and other behaviors that include timing, conditionals, conjunctions and disjunctions. 

We propose the use of STL because given a formula and state information from the robot and the environment, both semantic and performance metrics can be computed automatically, reducing the burden on the evaluator. The challenge is providing the state information; for physical robots we can use proprioceptive information and classifiers, as we show in Example~\ref{exm:bowlSTL}. For simulation evaluation, the world state is known, therefore STL metrics can shine.

\subsection{Performance metrics}
Ultimately, the purpose of learning  policies is to create autonomous robots; in many domains, robots will work with and around people. When working with people, a robot being ``correct" is not enough. The manner in which the robot behaves may impact the interaction and the acceptance of the robot~\cite{PercSafety,nonverbalHRI}. Here we suggest performance metrics for learned policies. 



\subsubsection{Signal Temporal Logic robustness}
\label{sec:STLRobust}
As mentioned above, STL has two types of semantics: Boolean and quantitative. Given an STL formula and a trajectory $X_t$, we can evaluate whether the trajectory evaluates to $True$ or $False$ (Boolean) but we can also calculate \textit{how close} the formula is to satisfying or violating the formula (robustness). A positive robustness indicates that the formula is $True$, negative that it is $False$. Thus, STL can be used to determine both correctness and quality of policies.
The definition of the qualitative semantics can be found in~\cite{donze2010robust}; in the following we use the RTAMT~\cite{nivckovic2020rtamt,rtamt} library to calculate the robustness of STL formulas on rollout data.

\begin{figure*}
     \centering
     \begin{subfigure}[b]{0.48\columnwidth}
         \centering        \includegraphics[width=\columnwidth]{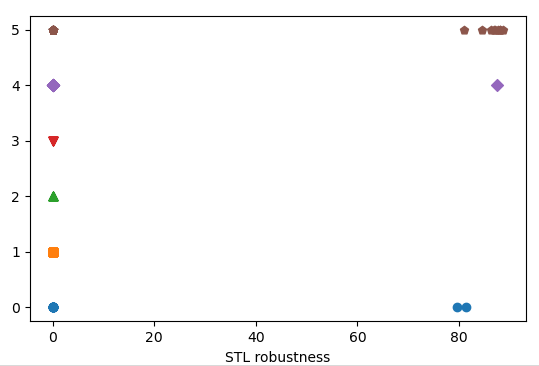}
         \caption{\textit{STL robustness}}
         \label{fig:STL}
     \end{subfigure}
     \begin{subfigure}[b]{0.48\columnwidth}
         \centering         \includegraphics[width=\columnwidth]{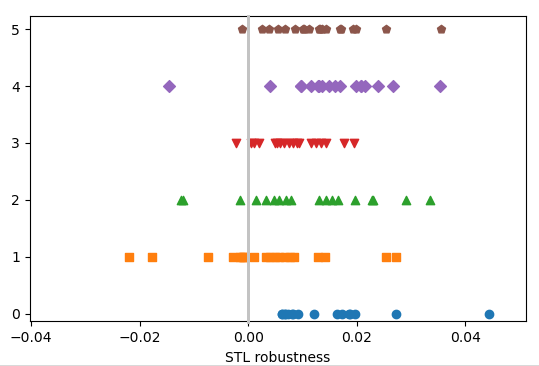}
         \caption{\textit{STL robustness zoomed in}}
         \label{fig:STLzoom}
     \end{subfigure}
        \caption{Robustness metric for Example~\ref{exm:bowlSTL} for the 6 policies (Y-axis). There are two types of behaviors; the points around 80 indicate that the robot did not make contact with the bowl, the points around 0 indicate that the robot did. In~\ref{fig:STLzoom} we can see which trajectories violated the STL formula and by how much - points with negative robustness represent rollouts for which during contact the end effector z-coordinate was smaller than 0.25. We added a gray line at zero to help visualize.}
        \label{fig:STLall}
\end{figure*}

\begin{example}[push bowl]
\label{exm:bowlSTL}
    Assume we prefer the robot not touch the table when it is pushing the bowl. To automatically evaluate this property we look at two signals; the $z$ coordinate of the robot end effector, as calculated by the Panda using forward kinematics (where the $z$ axis is perpendicular to the table, $z=0$ is the table and $z$ is positive for end effector positions above the table) , and $contact$, a signal that has values of around 20 for no contact and values greater than 500 when there is contact. We create this signal from a sensor attached to the end effector. The STL specification capturing the required property is $$ \Box((contact>100) \rightarrow (z > 0.25 )) $$ which means that at all times, whenever the robot is making contact, the value of $z$ must be greater than 0.25. Since this is an implication, the specification is satisfied if either the left hand side is $False$ (i.e. there is no contact), or the right hand side is $True$ (i.e. $z$ is always larger than 0.25) 

    Figure~\ref{fig:STLall} shows the robustness score of all the trajectories for the 6 policies. We can see two modes in these figures - trajectories with robustness of around 80 and trajectories with robustness around 0. The former corresponds to trajectories that do not make contact; the robustness corresponds to how close the left hand side of the implication is to being $False$. If the typical non-contact value is around 20, and our threshold is 100, we have 80 in robustness - $contact$ can increase by 80 before we need to satisfy the right hand side. 
    Trajectories around 0 correspond to trajectories in which the robot made contact. Points with negative values represent trajectories for which the minimum z value during contact was less than 0.25, thus violating the specification. 
    
    When comparing the policies, we can see that the brown pentagram (policy ID 5) is the most prone to not making contact at all (most runs with robustness $\sim 80$), while the orange square (policy ID 1) is the most prone to violating the $z$-value requirement (most runs with negative robustness).   
\end{example}
In our example we can see two types on behaviors because the signals themselves have a different range of values. This might not be the case in all situations, but in all cases the robustness will give insight regarding the success of a policy with respect to a possibly complex objective, and crucially, quantify how close the policy is to succeeding (or failing).

\subsubsection{Smoothness metrics}
Smoothness of the robot trajectory may impact the human-robot interaction~\cite{PercSafety}; Refai et. at.~\cite{MohamedRefai2021} explored several possible smoothness metrics and conclude that the most appropriate one is SPectral ARC length (SPARC)~\cite{Balasubramanian2015,sparcCode} computed over speed profiles. This metric is well suited for reaching tasks, and~\cite{Balasubramanian2015} discuss extensions to rhythmic movement. 
Looking at robot data, we can observe differences in the trajectory smoothness for different SPARC values; the more negative the value, the less smooth the trajectory is. 
 \begin{wrapfigure}{r}{0.5\textwidth}
    \begin{center}
  \includegraphics[width=0.48\textwidth]{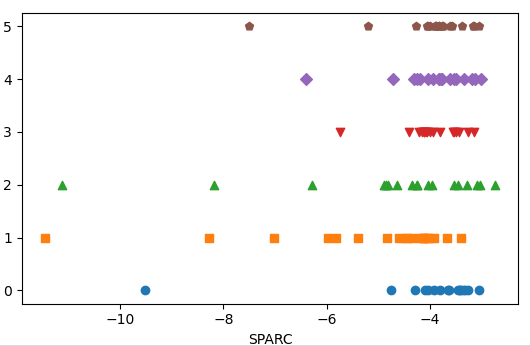}
  \end{center}
\caption{SPARC data for \textit{Bowl} (Example~\ref{exm:bowlSPARC}). We consider rollouts in which the robot makes contact with the bowl and we calculate SPARC for the pre-contact trajectory. 
Each data point corresponds to a rollout. Different policies are color coded. More negative SPARC value corresponds to less smooth motion.}
\vspace{-1cm}
  \label{fig:bowlSPARC}
\end{wrapfigure}

\begin{example}[Bowl SPARC]
\label{exm:bowlSPARC}
    Figure~\ref{fig:bowlSPARC} shows the SPARC values for robot end effector trajectories before the first contact for all the rollouts in which the robot made contact with the bowl, i.e. we removed rollouts in which the robot does not touch the bowl. 
    Looking, for example, at the policy represented by the green triangles (policy ID 2), we can see that the SPARC value ranges from -2.72 (smoothest) to -11.13 (least smooth). 
    The rollouts that correspond to the two extreme values are both successful; 
    however, the \emph{quality} of the motion is different with the smooth trajectory making contact with the bowl almost instantly and completing the task in 4 seconds while the least smooth moves in a periodic manner above the bowl before making contact (air balling) and taking 15 seconds to finish the task.  
\end{example}


\section{Analysis and Reporting}
The final aspect of our recommendations for best practices addresses the reporting of the results and the insights gained from the experiment to the community. We break down the information into three categories: experimental parameters, statistical analysis, and failures. We provide an example evaluation report in the appendix, Section~\ref{sec:evalExample}; in this evaluation we also demonstrate how we can gain insights regarding differences betwen policies despite them having similar success rates. 

\subsection{Experiment parameters}
We recommend that every experimental evaluation provide the following information: 1) a clear description of the semantic metrics, i.e. an explicit description of what is considered a success overall and in subgoals, 2) the number of evaluations performed across each condition and not just percentages (see Section~\ref{sec:stats}), 3) the timing of the evaluations, i.e. were all the policies evaluated in an A/B fashion, were they evaluated in one session, were evaluations performed across different days/weeks, and 
4) information regarding the initial conditions; visually as in Examples~\ref{exm:init} and ~\ref{exm:visInit} or as in~\cite{chi2024universal} (Fig.8), or in a narrative form. For the initial conditions we further recommend that researchers provide information regarding the relationship of the evaluation initial condition to the training initial conditions; were the evaluation ICs chosen as to try to capture in-distribution evaluation or were they chosen explicitly to evaluate out of distribution behavior?

\subsection{Statistical Analysis}
\label{sec:stats}
Providing a point estimate for success and performance metrics 
can be misleading~\cite{agarwal2021deep}. Conducting a statistical analysis provides the community with a more nuanced understanding of the results. 
The statistical analysis can follow the frequentist approach, by providing interval estimates~\cite{agarwal2021deep} or bounds~\cite{vincent2024generalizable}, or the Bayesian approach~\cite{Kruschke2013} of estimating parameters of distributions; in the following we illustrate the Bayesian approach and the importance of communicating the experimental details for the analysis. \cite{Kruschke2021} suggests best practices for reporting such statistical analysis.

\begin{example}[Bayesian Analysis]
\label{exm:BayesPancake}
We estimate the success rate of policies A and B for the \emph{Pancake} task. From Table~\ref{tab:pancake}, we see that the success rates of policy A and B are 83.3\% (15/18) and 64.7\% (11/17) respectively. In Bayesian analysis, we treat task  success as a Bernoulli distribution with (unknown) parameter $p$, the probability of success, which is a random variable we try to estimate given the data. Given a prior on $p$, here a uniform distribution between 0 and 1, and the data, we can estimate the distribution of $p$; the more the distributions of the policies overlap, the less confidence we have that one policy is better than the other. For this example, calculated with~\cite{Bayes}, we estimate a $.11$ probability that B actually performs better than A (Figure~\ref{fig:BayesPancake} in the appendix). 
\end{example}


Revisiting Example~\ref{exm:BayesPancake}, we illustrate why it is important to explicitly state the number of evaluations and not just the percentage. Keeping the percentage the same but changing the number of evaluations results in very different conclusions (as shown in Figure~\ref{fig:BayesPancakeNum} in the appendix). Reducing the number of evaluations makes any conclusions weaker, adding more evaluations makes them stronger. This is not surprising, but it emphasizes the need to provide additional information and analysis. 

\subsection{Failure modes}
We recommend providing a detailed description of common and surprising failure modes encountered during the evaluation. This includes information regarding failure categories, narrative descriptions, frequency, 
and visual descriptions through images and videos (e.g.~\cite{chi2024universal,liu2024okrobot}). 
This information is useful for several reasons; first, it sets expectations for researchers who are looking to incorporate learned models into their system as to what they can expect and what they need to be mindful of. Second, it provides a ``gradient" for researchers in robot learning regarding the current state of the art and where more research is needed. Third, it provides a baseline for future progress; demonstrated change in the type or frequency of failures corresponds to progress in the field.

\begin{example}[Failure description]
\label{exm:failBowl}
In the bowl example, we examine 6 different policies that were trained on  identical training data. One common failure mode is for the robot to miss the bowl when it approaches to establish contact. From the rubric and the STL robustness (Figure~\ref{fig:STLall}) we see that policy 5 is the most prone to this error with 9 failures, while policies 1,2, and 3 did not exhibit this failure during evaluation. This type of granular analysis can motivate additional evaluation or insight into architectural, pretraining, or other design decisions.  
\end{example}

\section{Discussion} 
\label{sec:discussion}

In the following we offer some additional thoughts regarding the empirical nature of robot learning.

\noindent \textbf{Purpose of using different metrics: }
Current learning techniques show a lot of promise for enabling robots to do tasks that were once considered impossible;
however, they are not yet sufficiently performant for real deployment. To close this gap, it is critical 
to track progress and identify areas of deficit that need to be addressed. 
Similarly to the development of nuanced metrics in computer vision, such an approach to robot policy evaluation will be critical to advance robotics as well.
Given the diversity of the approaches and tasks researchers consider, some metrics might make more sense than others. Furthermore, some metrics are going to be task specific, especially sub-goal achievement and STL specifications. 
We are advocating for researchers to choose the set of metrics that is best suited for their approach, task, and domain, but that provide more nuanced information than only success rate. 

\noindent \textbf{Simulation}: 
We focus on physical experiments; however, all the best practices discussed  are applicable to evaluation in simulation. Some aspects of the evaluation become straight forward, for example ensuring identical initial condition and environmental state (lighting, friction, etc) 
and automating all the metrics, at the expense of needing to reason about the sim-to-real gap. 
Closing this gap is an active area of research (e.g.~\cite{li2024evaluating}).

\noindent \textbf{Releasing evaluation data}:
The robot learning community has been active in publicly releasing training data and models (e.g.~\cite{open_x_embodiment_rt_x_2023}); however, while the data is well suited for ingestion by a machine learning algorithm, it is typically difficult to process manually. For example, it is not always clear what different arrays represent or how to decode images into videos that can be viewed by people. Furthermore, it is not common for researchers to publicly release evaluation data.   
While we propose several metrics, many useful metrics will be application dependent. For example automation applications may prefer approaches that exhibit consistency while human-interaction settings favor smoothness of motion. While we encourage researchers in the field to look beyond success rate in their evaluations, we also strongly advocate for open release of evaluation rollout data to enable additional nuanced posthoc analysis by the community.



\noindent \textbf{Closing thoughts:} Evaluating physical systems is difficult. It requires controlling numerous potentially confounding variables, balancing sample-size with diversity of conditions and comparisons given
a fixed resource budget, contending with equipment malfunction, and managing human subjectivity and bias. Despite these challenges, rigorous evaluation is absolutely critical for an empirically driven field, such as robot learning, to sustain lasting progress. In this work, we advocate for a set of best practices designed to mitigate the above issues. 
While not panacea, we believe widespread adoption of these practices will significantly reduce noise and in turn 
improve the pace of progress in the field.





\newpage
\section{Appendix}

\subsection{STL syntax and semantics}
 Signal temporal logic (STL) is defined over continuous-valued signals $X_t$ . The base element in the logic is a predicate $\mu$ and its associated predicate function $h(X)$. The truth value of $\mu$ at time $t$ is determined as: 
\begin{equation*}
\mu ::= 
    \begin{cases}
        True &  h(X_t) \geq 0.\\
        False &  h(X_t) < 0\\
    \end{cases}  
    \label{predicate}
\end{equation*}
In physical experiments, $X$ would typically be robot state, in simulation it can also include the state of environment such as object locations. 

An STL formula is recursively defines as~\cite{donze2010robust}:  $$\varphi := \mu~|~\neg\varphi~|~ \varphi_1 \wedge \varphi_2~ |~ \varphi_1\, \mathcal{U}_\mathcal{I}\, \varphi_2$$

\noindent where $\varphi, \varphi_1$ and $\varphi_2$ are STL formulas and $\mathcal{I}=[a,b]$ is an interval over which the formula is considered, where $0\leq a< b$. In practice, since we consider finite trajectories, $b < \inf$.

The logic contains Boolean operators: $\neg$ negation, $\wedge$ conjunction (``and"), and using those we can also construct $\vee$ disjunction (``or"), $\rightarrow$ implication, and $\leftrightarrow$ ``if and only if".
In addition, STL contains temporal operators: ``Until" $\mathcal{U}$ - the formula  $\varphi_1\,\mathcal{U}_\mathcal{I}\,\varphi_2$  evaluates to $True$ if $\varphi_1$ is $True$ until $\varphi_2$ becomes $True$ in interval $I$. Additional temporal operators are constructed using the Boolean operators and $\mathcal{U}$: ``Always'' $\Box_\mathcal{I}\varphi$ where $\varphi$ must be $True$ throughout the interval $\mathcal{I}$ for the formula to be $True$ and ``Eventually'' $\Diamond_\mathcal{I}\varphi$ where $\varphi$ must be $True$ at some point during the  interval $\mathcal{I}$ for the formula to be $True$. The semantics of STL can be found in~\cite{maler2004monitoring,donze2010robust}. Section~\ref{sec:STLRobust} provides a concrete example for such a metric; the sign of the robustness metric corresponds to the truth value of the Boolean semantics. 

\subsection{Visualizing initial conditions - additional example}

\begin{figure}[!h]{
  \centering
  \includegraphics[width=0.6\linewidth]{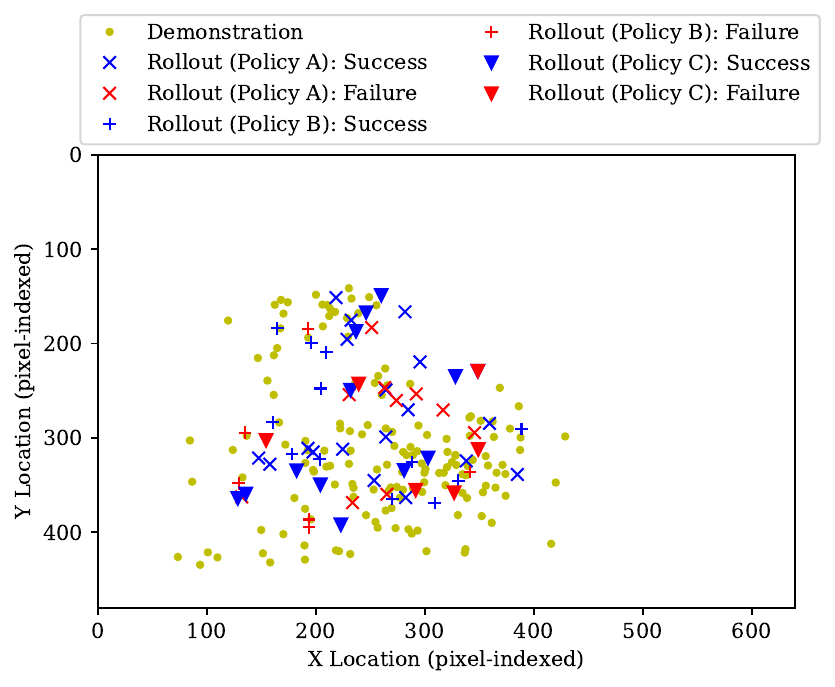}
  \caption{Initial location of the bowl in \textit{Bowl}}
  \label{fig:initialBowl}}
\end{figure}

\begin{table}
    \centering
    \begin{tabular}{ |p{4cm}||p{4cm}|  }
    \hline
    Number of demonstration & 154 \\
    \hline\hline
    Type of Policy & Success / Failure \\
    \hline
         Policy A & 18 / 10 \\
         Policy B & 12 / 6 \\
         Policy C & 13 / 6 \\
    \hline
    \end{tabular}
    \caption{Number of success and failure in \textit{Bowl} for the policies shown in Figure~\ref{fig:initialBowl}}
    \label{tab:bowl}
\end{table}

\begin{example}[Visualizing initial condition]
\label{exm:visInit}
    In Figure~\ref{fig:initialBowl} we visualize one aspect of the initial condition of a task - the initial location of the object that is to be manipulated. Figure~\ref{fig:initialBowl} shows the initial locations of the bowl for \textit{Bowl}. We used semantic segmentation to generate masks of the manipuland, then applied OpenCV~\cite{bradski2000opencv}'s SimpleBlobDetector to filter out misclassified pixels, leaving a single blob of maximum size per image, and finally calculated the 2D centroid of the manipuland pixels in the image frame. 

    In Figure~\ref{fig:initialBowl}, we compare the initial conditions of the evaluations of three policies. Table~\ref{tab:bowl} contains the respective success and failure numbers. While the success rate is similar, we can qualitatively see that the training distribution is not uniform and the evaluation conditions are not consistent between policies. 
\end{example}

\subsection{Bayesian analysis}
The following figures are the distributions of $p$ for Example~\ref{exm:BayesPancake}. They illustrate how success percentage alone, which is fixed for all the figures, does not allow for in depth analysis of the effectiveness of one policy with respect to another. 

\begin{figure}[!h]{
  \centering
  \includegraphics[width=0.7\linewidth]{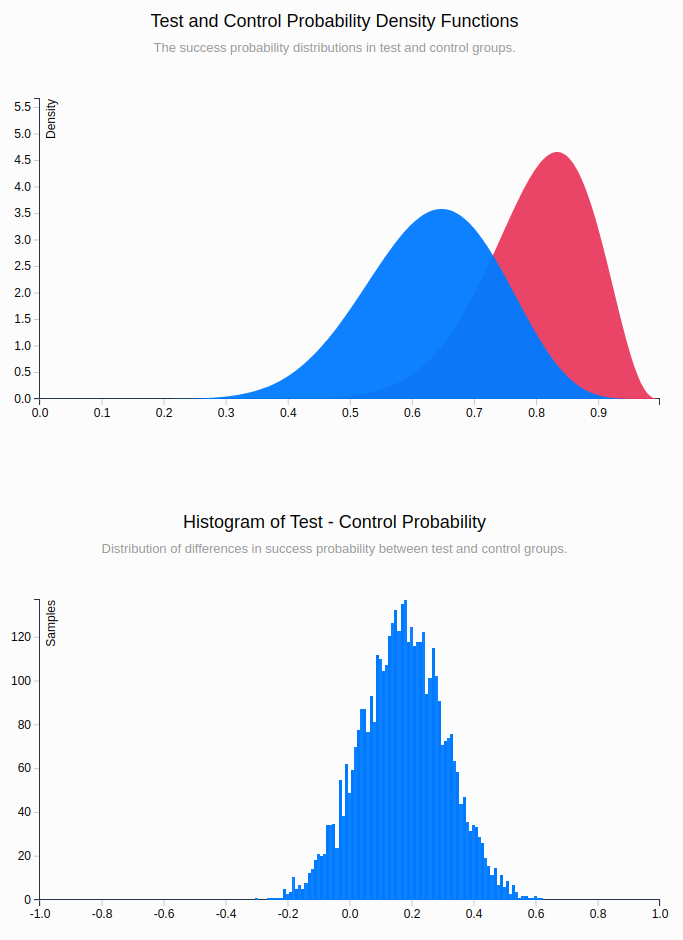}
  \caption{Estimated distributions of the success rate $p$ for the \textit{pancake} task (Example~\ref{exm:BayesPancake}). Red is policy A, blue is policy B. The top shows the estimated distributions, the bottom a histogram of the difference between $p$. We can see that a difference of 0 is in the support of the histogram. }
  \label{fig:BayesPancake}}
\end{figure}

\begin{figure*}
     \centering
     \begin{subfigure}[b]{0.48\columnwidth}
         \centering        \includegraphics[width=\columnwidth]{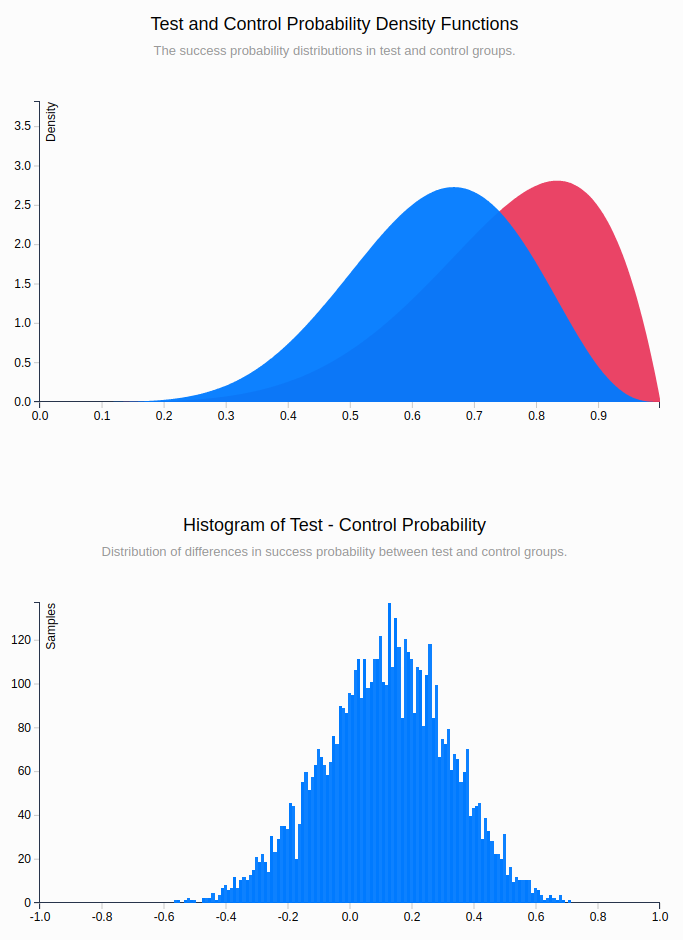}
         \caption{\textit{Less evaluations}}
         \label{fig:small}
     \end{subfigure}
     \begin{subfigure}[b]{0.48\columnwidth}
         \centering         \includegraphics[width=\columnwidth]{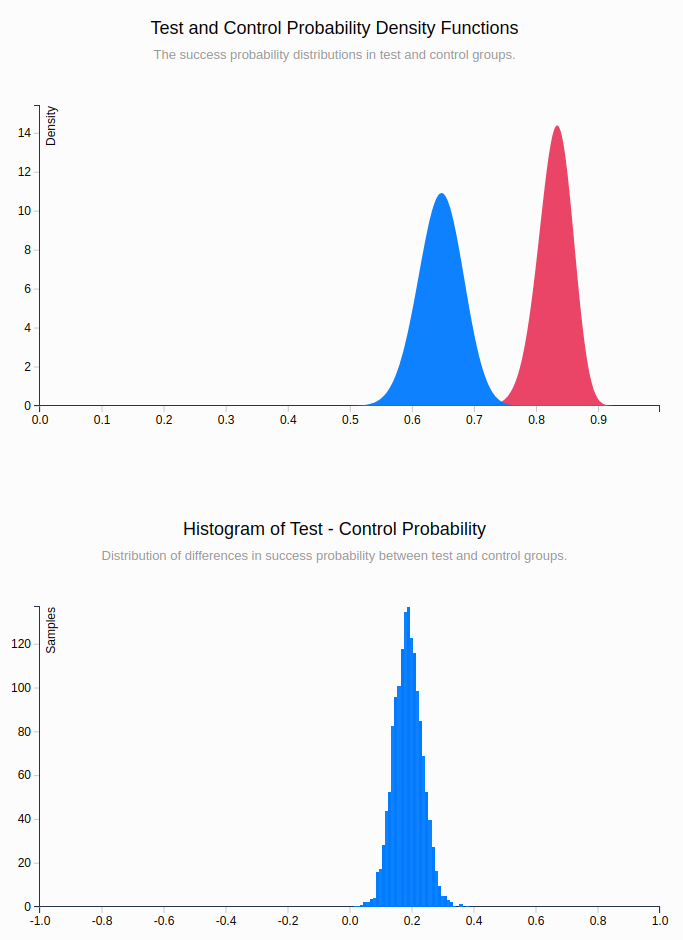}
         \caption{more evaluations}
         \label{fig:big}
     \end{subfigure}
        \caption{Estimated distributions of the success rate $p$ for the \textit{pancake} task (Example~\ref{exm:BayesPancake}) when we use the same success rate but vary the number of evaluations. Red is policy A, blue is policy B. The top shows the estimated distributions, the bottom a histogram of the difference between $p$. In Figure~\ref{fig:BayesPancake} policy success/failure is 15/3 for policy A and 11/6 for policy B. Here, Figure (a) is 5/1 for policy A and 6/3 for policy B (slightly different success rate because it must be natural numbers), and Figure (b) is 150/30 for policy A and 110/60 for policy B.  We can see that for (b), a difference of 0 between $p$ is no longer in the support of the histogram, indicating that A outperforms B under the experimental conditions. }
        \label{fig:BayesPancakeNum}
\end{figure*}

\subsection{Example Evaluation Report}
\label{sec:evalExample}
To illustrate the recommendations presented in this paper, we provide an example evaluation report for the comparison of two policies (policy A and policy B) learned through behavior cloning. The robot task is to pick up an energy bar and place it on a wooden tray. The robot is bimanual and can use either of its arms to perform the task. The details of the policies do not matter because we are not making a point about those policies but rather about how to report on their evaluation. If we were comparing the policies, in a discussion, we would draw conclusions regarding the policies based on the information presented below.

As shown below, even when we cannot draw conclusions regarding which policy is more successful, a detailed evaluation can lead to research insights and future directions.

\subsubsection{Experiment parameters}
\begin{figure}[!h]{
  \centering
  \includegraphics[width=0.8\linewidth]{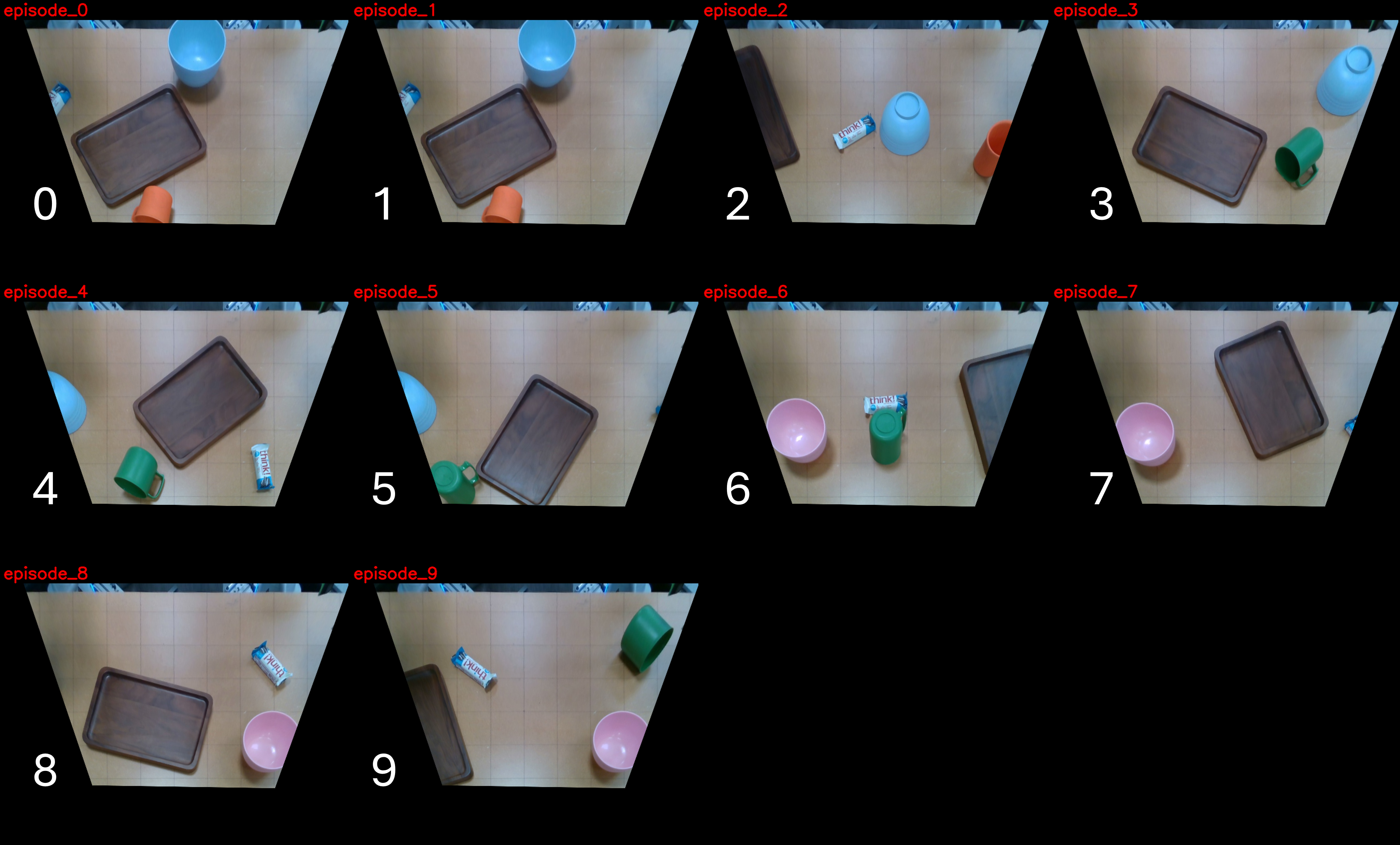}
  \caption{Initial location of the placing energy bar task}
  \label{fig:EnergyIC}}
\end{figure}

The policies were evaluated in an A/B fashion (blind to the evaluator) across two days, the first day each policy was run 12 times, the second day (4 days later) each policy was run 8 times. The evaluator was not involved in the data collection or policy training. We created 10 different initial conditions (ICs) that were similar to initial conditions in the training set and we evaluated each policy twice on each IC (for a total of 20 evaluations per policy) by visually matching the initial conditions using an image overlay tool. The ICs are shown in Figure~\ref{fig:EnergyIC}.

\noindent\textbf{Success criteria:} A run was considered successful if it ended with ``Energy bar is on the wooden tray and the tray in on the table." Note that we did not consider collisions with other items as failures.

\subsubsection{Results}
\noindent\textbf{Success rate}: Policy A succeeded in completing the task 13/20 times (0.65), while policy B succeeded 14/20 (0.7). Modeling the success of a policy as a binomial distribution, and assuming a uniform prior on the success probability, we cannot state that one policy outperforms the other. See Figure~\ref{fig:BaysEnergy} for the posterior distributions. 

Despite similar overall success, the policies performed differently and failed in different ways and on different ICs. 

\begin{figure*}
     \centering
     \begin{subfigure}[b]{0.48\columnwidth}
         \centering        \includegraphics[width=\columnwidth]{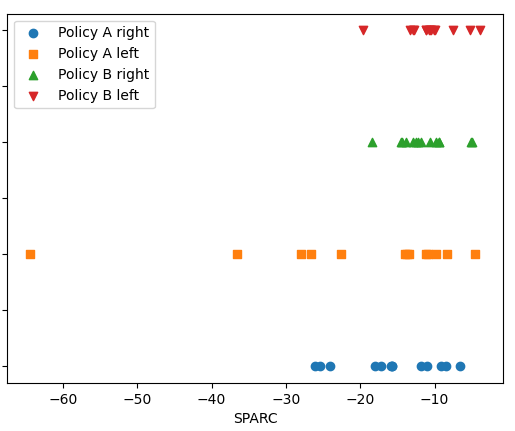}
         \caption{\textit{SPARC: smaller absolute value corresponds to smoother motion}}
         \label{fig:SPARCenergybar}
     \end{subfigure}
     \begin{subfigure}[b]{0.48\columnwidth}
         \centering         \includegraphics[width=\columnwidth]{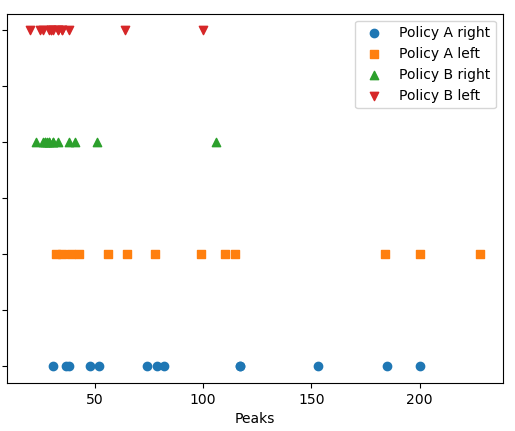}
         \caption{\textit{Velocity peaks: more peaks corresponds to longer and less smooth motions}}
         \label{fig:Peaksenergybar}
     \end{subfigure}
        \caption{Trajectory smoothness metrics for the example in Section~\ref{sec:evalExample} for the 2 policies and the two arms (Y-axis). }
        \label{fig:energybarSmoothness}
\end{figure*}

\begin{figure*}
     \centering
     \begin{subfigure}[b]{0.48\columnwidth}
         \centering        \includegraphics[width=\columnwidth]{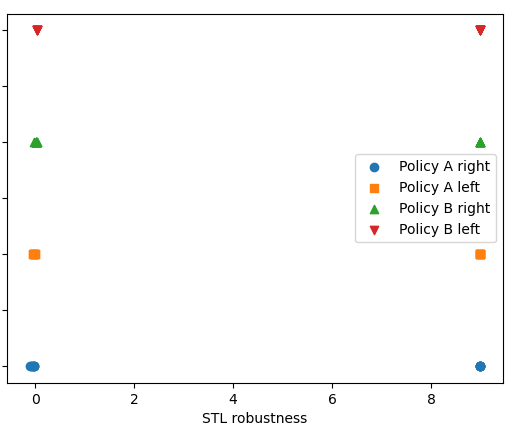}
         \caption{\textit{STL robustness}}
         \label{fig:STLenergybar}
     \end{subfigure}
     \begin{subfigure}[b]{0.48\columnwidth}
         \centering         \includegraphics[width=\columnwidth]{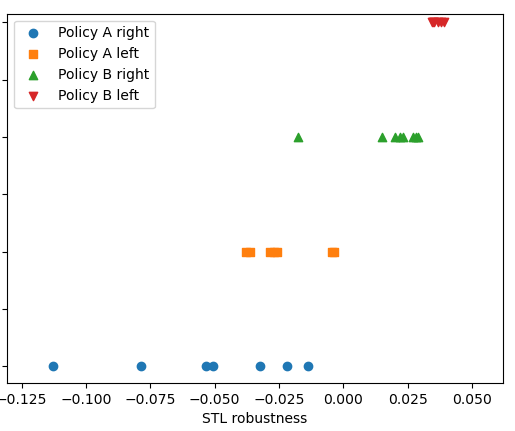}
         \caption{\textit{STL robustness zoomed in}}
         \label{fig:STLzoomenergybar}
     \end{subfigure}
        \caption{STL Robustness metric for the example in Section~\ref{sec:evalExample} for the 2 policies and the two arms (Y-axis). There are two types of behaviors; the points around 9 indicate that the robot did not close its gripper, the points around 0 indicate that the robot did. In~\ref{fig:STLzoomenergybar} we can see which trajectories violated the STL formula and by how much - points with negative robustness represent rollouts for which when the gripper was closing, the z-coordinate was larger than 0.25 indicating that the robot was attempting to grasp too high. }
        \label{fig:STLallenergybar}
\end{figure*}

\begin{figure}[!h]{
  \centering
  \includegraphics[width=0.6\linewidth]{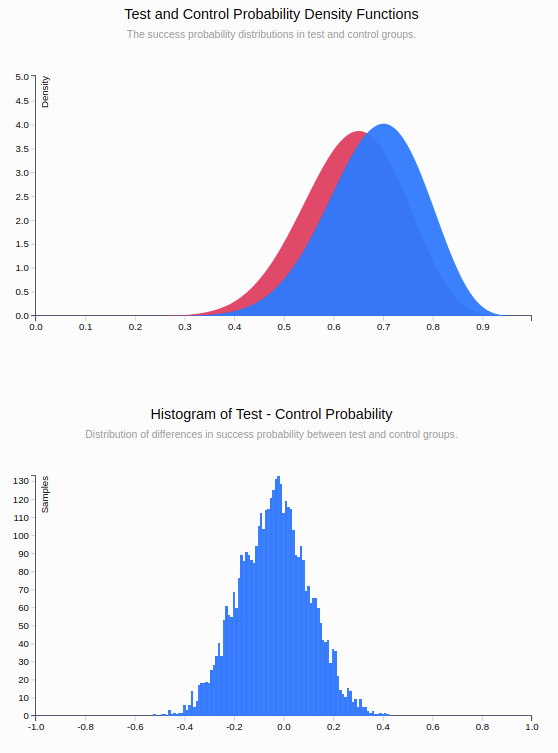}
  \caption{Estimated distributions of the success rate $p$ for the energy bar task, calculated with~\cite{Bayes}. Red is policy A, blue is policy B. The top shows the estimated distributions, the bottom a histogram of the difference between $p$. We can see that a difference of 0 is in the support of the histogram. }
  \label{fig:BaysEnergy}}
\end{figure}

\noindent \textbf{Performance:} The following performance analysis contains only the \textit{successful runs} for each of the policies. In Figures~\ref{fig:energybarSmoothness} and \ref{fig:STLallenergybar} we show the performance metrics for both policies and for both robot arms.

\textbf{\textit{Smoothness of the trajectories:}} Figure~\ref{fig:energybarSmoothness} shows two metrics that relate to the smoothness of the trajectory. Figure~\ref{fig:SPARCenergybar} show the SPARC metric--smaller absolute values correspond to smoother trajectories. Figure~\ref{fig:Peaksenergybar} shows the number of peaks in the velocity of the end effector; more peaks correspond to longer and less smooth trajectories.

From Figure~\ref{fig:energybarSmoothness}, we can see that Policy B (triangles) produces smoother trajectories according to both smoothness metrics.

\textbf{\textit{STL robustness example:}} We observed qualitatively that policy B's grasps of the energy bar are more stable than those of policy A. To quantify this, we write an STL specification over two signals: $z$, the z-value of the end effector in a global frame centered on the table, and $gripper\_diff$ which is the difference in gripper width between two consecutive time steps; when the gripper is closing, $gripper\_diff > 0$ and when it is opening $gripper\_diff < 0$ . We plot the robustness of the following STL formula: 
$$ \Box((gripper\_diff*1000 > 9) \rightarrow (z < 0.25 ))$$
This formula states that always, if the difference between the gripper width in two consecutive time steps is greater than 0.009 (an observed change when the gripper is closing) then the end effector z-value should be less than 0.25. This formula is true and has positive robustness metric if either 1) the robot does not close its gripper (the left side of the implication is false) or 2) the end effector is close to the table when the gripper closes.

Figure~\ref{fig:STLallenergybar} shows the robustness metric for both policies and both arms. We write the STL formula as $gripper\_diff*1000 > 9$ and not as $gripper\_diff > 0.009$ so that we can observe the different modes in the robustness metric (gripper did not close resulting in a robustness of 9 vs gripper closed and a resulting robustness depending on the value of $z$). From the figure we can see that each policy and each arm had runs in which the gripper did not close. That is expected because each run had only one arm that grasped the energy bar. In the zoomed in figure (Figure~\ref{fig:STLzoomenergybar}) we see that policy B (triangles) has higher robustness values, meaning that policy B tends to grasp lower than policy A. This is consistent with the failure modes we observed as described below.

\noindent\textbf{Failure analysis:} Both policies failed on IC 6; Policy A also failed on both runs of IC 4 and one run each of ICs 7,8,9, while policy B failed on both runs of ICs 2 and 9. 

Policy A's main failure mode was the grasping of the energy bar, it either failed to grasp or dropped it prematurely (6/7 failures). Policy B, on the other hand, consistently picked up the energy bar (20/20 runs) but then either moved away from the tray or placed the energy bar in a different location (6/6 failures). 

These different failures modalities may have different implications in a human-robot interaction scenario, so while the success rate is similar, one policy might be considered better by users of the robot, if this information is provided.  Furthermore, for robot learning researchers, this information may provide insight on the algorithmic choices and future research directions.

\bibliography{references}

\begin{thebibliography}{51}
\bibitem[say()]{saycan2022arxiv}
Do as i can and not as i say: Grounding language in robotic affordances.
\end{thebibliography}

\end{document}